\documentclass{article}

\usepackage{microtype}
\usepackage{graphicx}
\usepackage{subfigure}
\usepackage{booktabs} 

\usepackage{amsmath,amssymb}
\usepackage{amsthm}
\usepackage{multirow}
\usepackage[table,xcdraw]{xcolor}
\usepackage{kotex}
\usepackage{enumitem}

\newtheorem{definition}{Definition}

\usepackage{hyperref}

\usepackage[accepted]{icml2021}

\icmltitlerunning{Unsupervised Embedding Adaptation via Early-Stage Feature Reconstruction}

\begin{document}

\twocolumn[
\icmltitle{Unsupervised Embedding Adaptation via Early-Stage Feature Reconstruction for Few-Shot Classification}

\begin{icmlauthorlist}
\icmlauthor{Dong Hoon Lee}{KAIST}
\icmlauthor{Sae-Young Chung}{KAIST}
\end{icmlauthorlist}

\icmlaffiliation{KAIST}{School of Electrical Engineering, Korea Advanced Institute of Science and Technology (KAIST), Daejeon, Korea}

\icmlcorrespondingauthor{Dong Hoon Lee}{donghoonlee@kaist.ac.kr}

\icmlkeywords{few-shot classification, unsupervised learning}

\vskip 0.3in
]

\printAffiliationsAndNotice{}  

\begin{abstract}
We propose unsupervised embedding adaptation for the downstream few-shot classification task.
Based on findings that deep neural networks learn to generalize before memorizing, we develop Early-Stage Feature Reconstruction (ESFR) --- a novel adaptation scheme with feature reconstruction and dimensionality-driven early stopping that finds generalizable features.
Incorporating ESFR consistently improves the performance of baseline methods on all standard settings, including the recently proposed transductive method.
ESFR used in conjunction with the transductive method further achieves state-of-the-art performance on \textit{mini}-ImageNet, \textit{tiered}-ImageNet, and CUB; especially with 1.2\%$\sim$2.0\% improvements in accuracy over the previous best performing method on 1-shot setting.\footnote{Code is available at https://github.com/movinghoon/ESFR}
\end{abstract}
 \section{Introduction}
Deep learning has achieved impressive results on visual recognition tasks.
However, it still has difficulty generalizing to novel classes with few examples; while humans can learn to recognize from few experiences.
Few-shot classification \cite{miller2000learning, vinyals2016matching, ravi2016optimization} is designed to bridge the gap between the two and has recently attracted substantial attention.

Several works \cite{TPN, CAN, Team, SIB, transBaseline, LaplacianShot, TIM} have shown the effectiveness of transductive methods in few-shot classification, showing a significant improvement over inductive methods.
While test samples are inaccessible in an inductive few-shot classification setting, one can utilize all the unlabeled test samples together to make an inference in a transductive setting.
The co-existence of labeled- and unlabeled-data in this setting motivates the use of transductive inference or semi-supervised learning.
A popular transductive approach is pseudo-label-based methods that progressively update the labels or inference models by predicted test samples.
For instance, \citet{TPN, kim2019edge} use a neighbor graph for label propagation, \citet{CAN, BDCSPN} use predicted labels on test samples to update the class prototypes, and \citet{SCA, SIB} use prediction to produce an intrinsic loss or synthetic gradient.
Another line of works utilizes regularization terms on unlabeled test samples. 
To list a few, \citet{transBaseline, TIM} use entropy minimization of the prediction on unlabeled data, \citet{LaplacianShot} uses the Laplacian regularization term for graph clustering.
These methods are mostly originated from semi-supervised learning studies; approaches in semi-supervised learning are still motivating few-shot classification research.

On the other hand, semi-supervised learning research has recently benefited from unsupervised learning.
Rapidly advancing self-supervised learning methods \cite{caron2021unsupervised, grill2020bootstrap, chen2020big} have shown strong performance on semi-supervised image classification tasks.
A popular approach is to use a self-supervision loss for representation learning to acquire more general features.
Learned representations are then used with fine-tuning \cite{caron2021unsupervised, grill2020bootstrap, chen2020big} or other semi-supervised learning methods \cite{zhai2019s4l, kim2021selfmatch} for downstream tasks.
These studies achieved state-of-the-art performance on semi-supervised learning tasks, especially in settings with extremely few labels.

The success of unsupervised learning on semi-supervised tasks suggests the potential benefit of finding shared features or patterns without labels in relevant research areas. In few-shot classification, several works \cite{Gidaris_2019_ICCV, self_eccv} use additional self-supervision loss during the training of base datasets to learn more general features. However, the use of unsupervised learning on unlabeled data that appears in test-time is less studied. In this work, we study unsupervised learning for adaptation to satisfy the thirst.
\newpage
Our contributions are summarized as follows:
\begin{itemize}
	\item We find that early generalized features during unsupervised training are valuable for recognizing novel classes of few-shot classification. Based on recent studies of deep neural network's training dynamics, we explain the finding with experiments.
	(Section~\ref{subsection:feature_reconstruction})
	
	\item Based on the finding, we construct a novel embedding adaptation scheme with (1) feature reconstruction training and (2) dimensionality-driven early stopping. Our method provides task-adapted embeddings composed of desirable-shared features, which are more likely to be task-relevant and valuable for the few-shot classification. Our method is used as a plug-and-play module for few-shot methods.
	(Section~\ref{subsection:proposed})
	
	\item We test our method, ESFR, used in conjunction with baseline methods in the standard few-shot classification benchmarks. ESFR consistently improves the performance of baselines; adding ESFR to the transductive method achieves the state-of-the-art performance on \textit{mini}-ImageNet, \textit{tiered}-ImageNet, and CUB.
	Particularly in the scarce-label setting of 1-shot, our method outperforms the previous state-of-the-art with accuracies of 1.2\%$\sim$2.0\%. (Section~\ref{subsection:main_results})
\end{itemize} \section{Preliminaries}
\subsection{Problem Setting}
In a few-shot classification task, a small labeled support set $S=\{(x^i_s,y^i_s)\}_{i=1}^{|S|}$ and unlabeled query set $Q=\{(x^i_q,)\}_{i=1}^{|Q|}$ are given.\footnote{$(x,y)$ denotes an image and its label.}
Both support samples and query samples are from the same novel classes that are never seen during training.
The goal of few-shot classification is to classify query (test) samples by few examples in the support set.
In the usual setting, the support set has $K=1$ or $5$ examples per $N=5$ novel classes, and we call this an $N$-way $K$-shot problem.

We address a transductive few-shot classification task where all query samples are accessible.
Since learning from few samples without prior is extremely hard, we use a pre-trained embedding network $f$ that is trained on the base dataset as in \citet{LEO, SimpleShot, SIB, LaplacianShot}.
We denote $S_f=\{(f(x^i_s),y^i_s)\}_{i=1}^{|S|}$ and $Q_f=\{(f(x^i_q),\cdot)\}_{i=1}^{|Q|}$ as the support set and the query set in the embedding domain, respectively.

Our interest is in task-adapted embeddings (representations) that are useful in the given few-shot task.
We construct the embeddings with a module $g_\phi$ on top of $f$ by training on the union of the support set and query set.

\subsection{Local Intrinsic Dimensionality (LID)}
\label{subsection:LID}
We briefly explain LID that our method uses as early stopping criterion.
LID is a statistical version of an expansion-based Intrinsic Dimension (ID) that provides an estimated subspace dimension of local regions.
Recently, substantial redundant dimensions of modern deep neural networks lead to the wide use of ID and LID to track and analyze training \cite{Amsaleg17, Ma18a, Ma18b, Ansuini19, Gong19}.
The formal definition of LID is given as \cite{Amsaleg15, Houle17a, Houle17b}:
\begin{definition}[Local Intrinsic Dimensionality]
	Given a data point $x$, let $r>0$ be a continuous random distance variable from $x$. For the cumulative density function $F_x(r)$, the {\normalfont LID} of $x$ at distance r is defined as{\normalfont:}
	\begin{equation}
	\text{\normalfont LID}(r;F_x) \overset{\underset{\mathrm{def}}{}}{=} \lim_{\epsilon\rightarrow 0^{+}}\frac{\ln{F_x((1+\epsilon)r)} - \ln{F_x(r)}}{\ln(1+\epsilon)}, \end{equation}
	whenever the limit exists. The {\normalfont LID} at $x$ is then defined as the limit of distance $r\rightarrow 0^{+}${\normalfont:}
	\begin{equation}
	\text{\normalfont LID}(x) \overset{\underset{\mathrm{def}}{}}{=} \lim_{r\rightarrow 0^{+}} \text{\normalfont LID}(r;F_x).
	\end{equation} 
\end{definition}

Since the density function of the distance variable is usually unknown, the exact value of LID is hard to acquire.
We use the maximum likelihood estimation by \citet{Amsaleg15} to calculate the LID estimates as follows:
\begin{equation}
\widehat{\text{LID}}(x) = -\left[\frac{1}{m}\sum_{i=1}^m \ln\frac{r_i(x)}{r_m(x)}\right]^{-1},
\end{equation} 
where $r_i(x)$ indicates the distance\footnote{We use Euclidean distance.} between $x$ and its $i$-th nearest neighbor.
The number of the nearest neighbor $m$ should be chosen appropriately to make estimation local but stabilized.\footnote{We set $m=20$ throughout experiments as in \citet{Ma18a}.}

We refer to \citet{Amsaleg15, Houle17a, Houle17b} for more details about LID and its estimation methods.   \begin{figure*}[ht]
	\vskip -0.1in
	\begin{center}
		\centerline{\includegraphics[width=16cm]{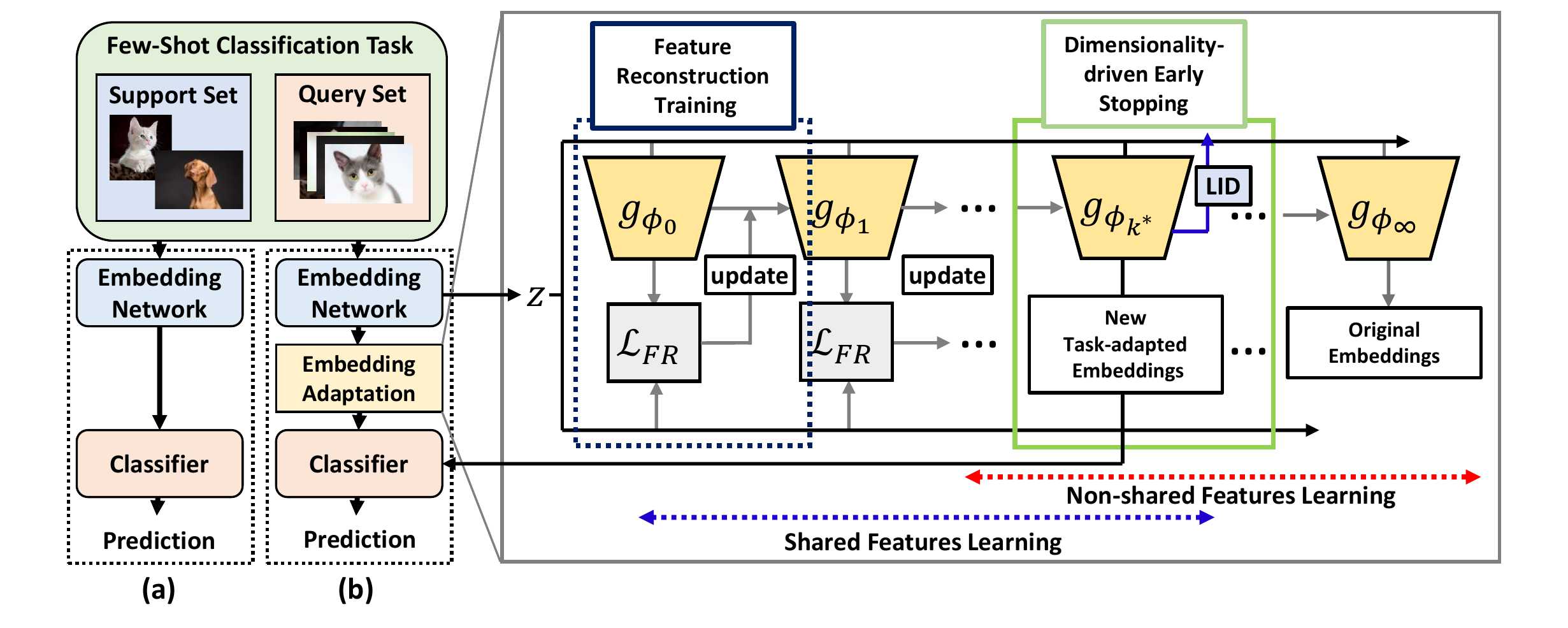}}
		\vskip -0.15in
		\caption{
			Overview of our method.
			1-(a) shows the case without embedding adaptation, and 1-(b) shows the case with embedding adaptation.
			Our scheme mainly consists of \textit{feature reconstruction training} and \textit{dimensionality-driven early stopping}, and provides new embeddings of generalizable features for the downstream few-shot task.
	}
		\label{Overview}
	\end{center}
	\vskip -0.3in
\end{figure*}
\section{Methodology}
Figure~\ref{Overview} illustrates the usage and overview of our method.
As a plug-and-play module, our method provides task-adapted embeddings to other few-shot classification methods.
Our method is mainly composed of feature reconstruction training and LID-based early stopping.
We will explain feature reconstruction training (Section~\ref{subsection:feature_reconstruction}); LID-based early stopping (Section~\ref{subsection:early_stopping}); the overall method (Section~\ref{subsection:proposed}), in the following.

\subsection{Feature Reconstruction}
\label{subsection:feature_reconstruction}
Our main idea is based on the findings that \textit{deep neural networks learn to generalize before memorizing} to abstract task-useful features. We design feature-level reconstruction training, which is unsupervised learning and builds on prior knowledge given by embeddings. As mentioned before, unsupervised learning for few-shot adaptation tends to have less attention; and we find that naively applied unsupervised learning for few-shot adaptation often fails.\footnote{We test self-supervised learning models of  rotation \cite{Rotnet} and jigsaw \cite{jigsaw} in Appendix E.} Moreover, the behavior of unsupervised learning with a pre-trained embedding network is often ambiguous. For instance, contrastive learning heavily relies on augmentations, while augmentations affect embeddings differently depending on how the embedding network is pre-trained. Instead, we find that our feature reconstruction training can be used to adapt embeddings for few-shot classification.

We explain our feature reconstruction training.
For a few-shot classification task with embedding support set $S_f$ and query set $Q_f$; we train a reconstruction module $g_\phi$ using the following feature level reconstruction loss:
\begin{equation}
\mathcal{L}(\phi) = \frac{1}{|S_f\cup Q_f|}\sum_{z\in S_f\cup Q_f} d_\text{cos}(z, g_\phi(z)),
\end{equation}
where $d_\text{cos}$ denotes the cosine distance. Both $z$ and $g_\phi(z)$ are preprocessed\footnote{Details are described in the experiment section.} embeddings, but their expressions are omitted for notational simplicity. We note that for a newly given few-shot classification task, the weight $\phi$ of the reconstruction module is randomly re-initialized.

\begin{figure}[t]
	\vskip -0.1in
	\begin{center}
		\centerline{\includegraphics[width=\columnwidth]{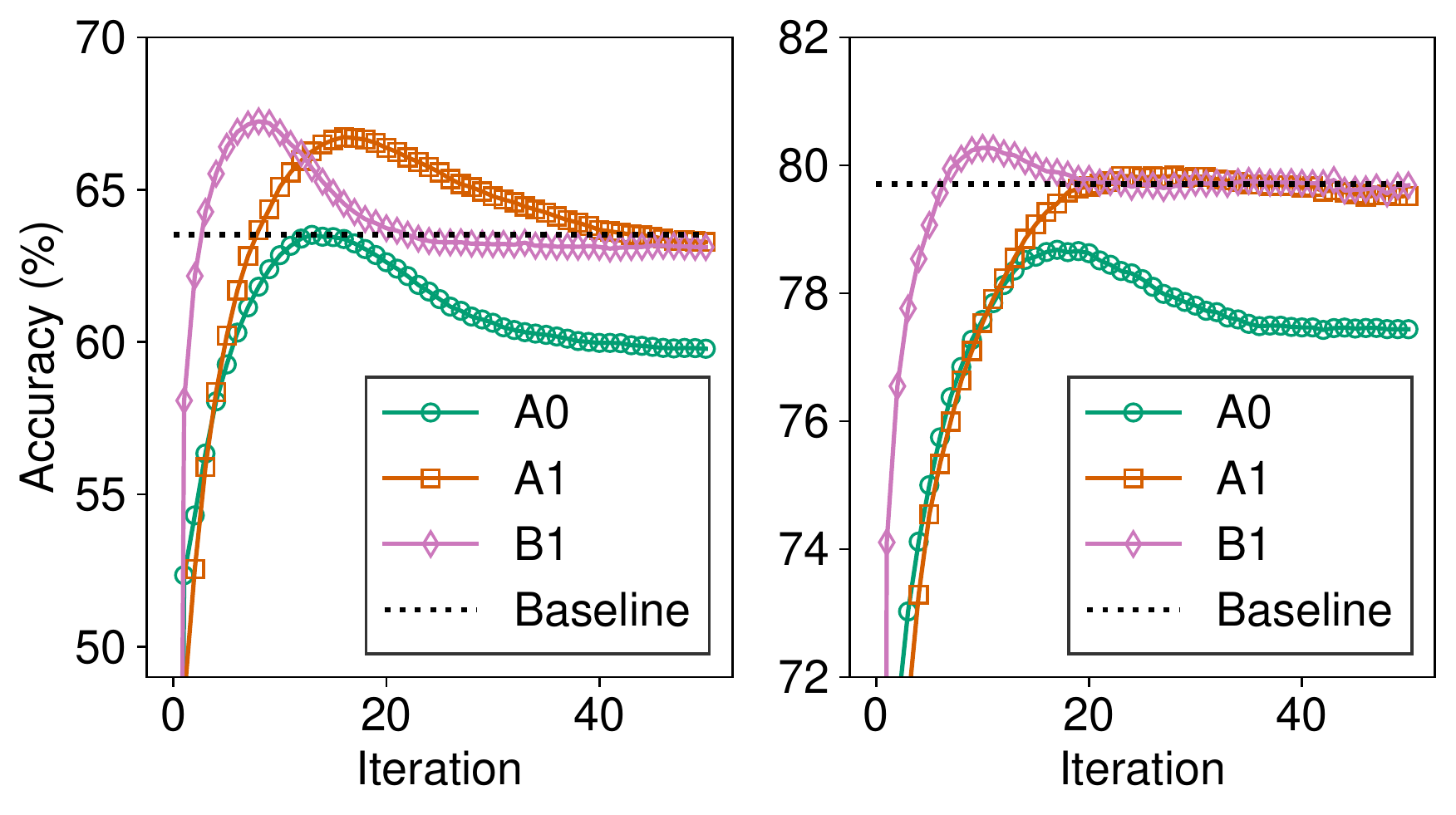}}
		\vskip -0.15in
		\caption{
			Few-shot classification accuracy with \textbf{A0}, \textbf{A1}, and \textbf{B1} during feature reconstruction training. 
			We use ResNet-18 backbone, \textit{mini}-ImageNet dataset, and nearest neighbor classifier.
			For $g_{\phi_1}$, we use a 4-layer neural network with 256-128-256-512 units.
			Other settings are described in the experiment section.
			(left) shows the accuracy on 1-shot and (right) shows the accuracy on 5-shot setting.}
		\label{graph:training_dynamics}
	\end{center}
	\vskip -0.4in
\end{figure}
We investigate the behavior of reconstruction modules during feature reconstruction training \textit{w.r.t.} the downstream few-shot task. We train two types of reconstruction modules, $g_{\phi_1}$ with compressed (bottleneck) hidden layer as in conventional auto-encoders, and $g_{\phi_2}$; without compression. We evaluate few-shot classification accuracy with three embeddings \textbf{A0}, \textbf{A1}, \textbf{B1}; where \textbf{A0} is the middle compressed hidden layer output of $g_{\phi_1}$; \textbf{A1} and \textbf{B1} are the reconstructed output of $g_{\phi_1}$ and $g_{\phi_2}$, respectively.

Figure~\ref{graph:training_dynamics} shows an interesting behavior that few-shot classification accuracies, with embeddings of reconstruction modules, initially increase then decrease. Moreover, the peak accuracy of \textbf{B1} exceeds the baseline accuracy of the original embedding, on both 1- and 5-shot settings. 
We believe that recent studies of Deep Neural Networks (DNNs) explain such behavior. Several works \cite{ArpitL17, Lampinen19, stephenson21} observe a property that DNNs learn to generalize before memorizing. To be more specific, \citet{ArpitL17} reports that DNNs learn patterns first before memorization, with experiments on the mixture of well-structured real data and noisy data. Further research \citet{Lampinen19, stephenson21} provide analytical explanations on the property that learning speeds between generalization of patterns and memorization of noise are different.\footnote{Generalization is faster than memorization.} We argue that the behavior, shown in Figure~\ref{graph:training_dynamics}, is a result of the DNNs' property. By the property, reconstruction modules learn shared features faster since they form certain patterns or correlations among data, while non-shared features are learned later since they are less generalizable. Generalizable shared features are more likely to be task-relevant in classification; the difference in learning speeds between shared features and non-shared features explains the initial increase of accuracies.

Our main idea is to use the behavior shown in Figure~\ref{graph:training_dynamics} for embedding adaptation.
The behavior provides an opportunity to achieve improved few-shot classification performance by acquiring new embeddings of generalized features.
To do this, we find that non-compressed and non-encoded embedding (\textbf{B1}) performs the best. Throughout the rest of our work, we use the reconstructed output $g_\phi$ that is non-compressed as task-adapted embeddings.

\begin{figure}[ht]
	\vskip -0.1in
	\begin{center}
		\centerline{\includegraphics[width=\columnwidth]{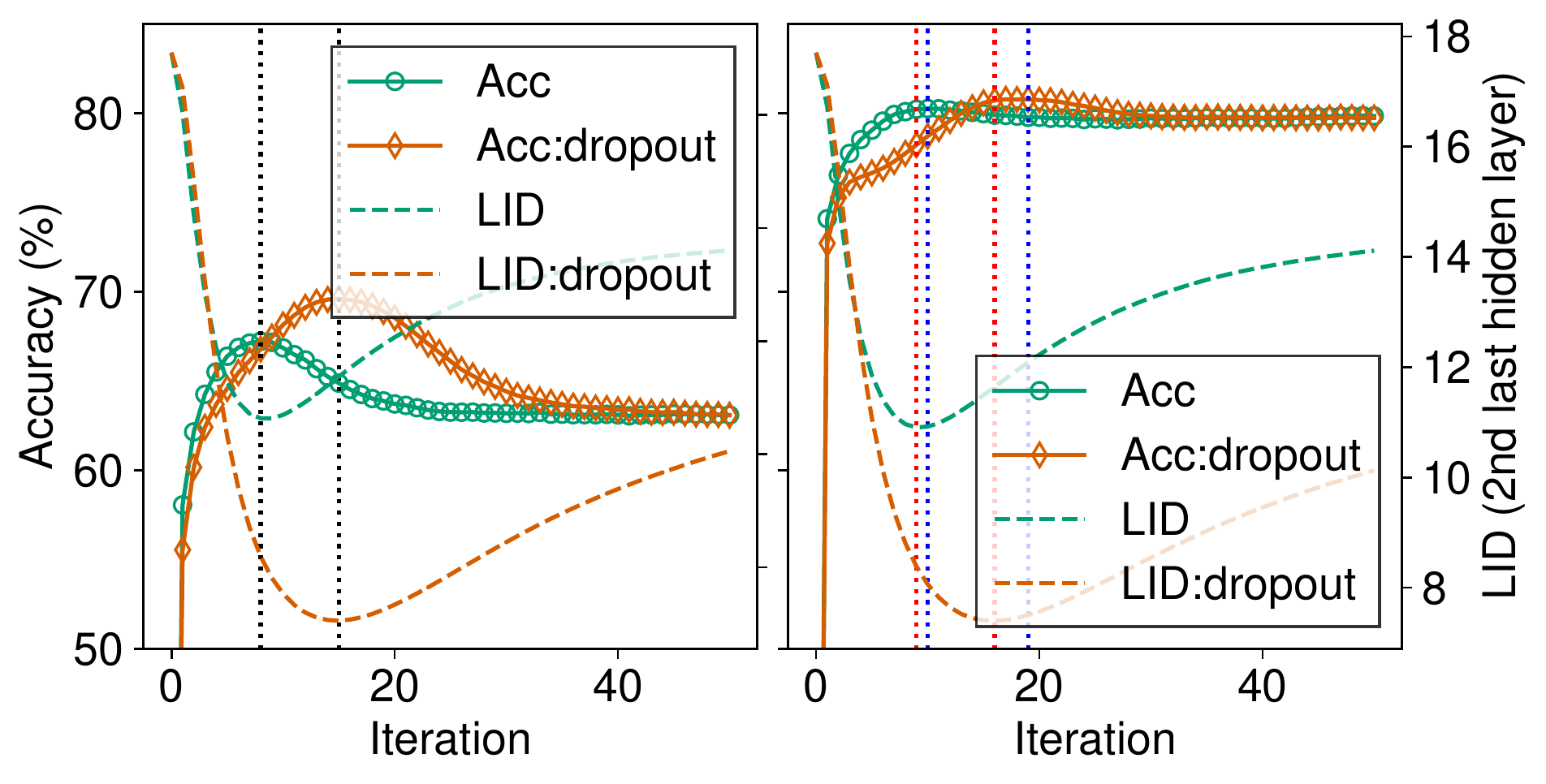}}
		\vskip -0.15in
		\caption{
			The figure shows the accuracy and the LID curve during feature reconstruction training, with (Acc:dropout, LID:dropout) and without (Acc, LID) dropout perturbation.
			We use the same setting as in Figure~\ref{graph:training_dynamics}.
			The vertical lines indicate the maximum accuracy (blue) and the minimum LID (red) points; for 1-shot, they are overlapped.
			(left) shows the results in the 1-shot and (right) shows the results in the 5-shot setting.
		}
		\label{graph:lids}
	\end{center}
	\vskip -0.25in
\end{figure}
We propose to perturb the training to further discard the non-shared features.  Perturbation with noise does not form generalizable patterns or correlations; thus, perturbation tends to make training harder for less generalizable non-shared features while shared features are still learnable by pattern learning. While additive perturbation is hard to design, due to the unknown distribution of $z$, we find that multiplicative perturbation by dropout \cite{dropout} is well suited for our purpose. Our feature reconstruction training with input dropout uses the following loss function\footnote{$\circ$ indicates the element-wise product.}:
\begin{equation}
\label{equation:loss}
\mathcal{L}_{\text{FR}}(\phi) = \frac{1}{|S_f\cup Q_f|}\sum_{z\in S_f\cup Q_f}\mathbb{E}_\mu [d_\text{cos}(z, g_\phi(z \circ \mu))],
\end{equation}
where $\mu$ is a multiplicative noise implemented with dropout.
Figure~\ref{graph:lids} shows the effect of dropout on the training curve; we can observe that the dropout perturbation results in higher peak accuracies. \subsection{Dimensionality Driven Early Stopping}
\label{subsection:early_stopping}
Utilization of early retained generalizable features seems a promising idea for the downstream few-shot classification. However, determining the optimal early stopping time is not straightforward. Here, we suggest using Local Intrinsic Dimensionality (LID) as an early stopping criterion. Recent work by \citet{Ma18b} used LID to monitor the internal generalization and memorization during training.
They argued that LID tends to decrease while DNNs learn to generalize, due to summarizing effect; LID tends to increase while memorizing, as DNNs try to encode detailed information individually.
We observe the tendency in our reconstruction training, and we find that LID can be used as an early stopping criterion.

For a given few-shot classification task with embedding support set $S_f$ and query set $Q_f$, we use estimated LID, with reconstruction module $g_{\phi}$, given as:
\begin{align}
\label{equation:lid}
\widehat{\text{LID}}(\phi) &= \sum_{z\in S_f\cup Q_f} \widehat{\text{LID}}(g^{L-2}_\phi(z)) \notag \\
&= -\sum_{z\in S_f\cup Q_f} \left[\frac{1}{m}\sum_{i=1}^m \ln\frac{r_i(g^{L-2}_\phi(z))}{r_m(g^{L-2}_\phi(z))}\right]^{-1},
\end{align}
where $g^{L-2}_\phi$ is the hidden representation of the second-to-last layer of $g_\phi$; $r_i(g^{L-2}_\phi(z))$ denotes the Eucidean distance between $g^{L-2}_\phi(z)$ and its $i$-th nearest neighbor in $\{g^{L-2}_\phi(z')|z'\in S_f\cup Q_f\}$.
Our $\widehat{\text{LID}}(\phi)$ is a proxy for the hidden layer subspace dimensionality of the reconstruction module.

In Figure~\ref{graph:lids}, we empirically investigate the relationship between the LID and accuracy during reconstruction training.
The result shows that the change of LID can be used to find the early stopping time of the best possible new embeddings; when the LID becomes the lowest or starts to increase.
To be more specific, in 1-shot settings, we observe that the LID behaves exactly the opposite of the accuracy curve, for both with and without dropout.
For 5-shot settings, the LID behaves almost the opposite of the accuracy curve; however, it has a small misalignment, especially for the case with dropout.
This seems reasonable since the classifier can further filter out non-shared features by multiple examples of novel classes in 5-shot settings; hence acquiring more shared features at the cost of learning non-shared ones can be advantageous. \begin{algorithm}[tb]
	\caption{ESFR}
	\label{alg:proposed}
	\begin{algorithmic}
		\STATE {\bfseries Input:} embedding support set $S_f$, embedding query set $Q_f$, and few-shot classifier $\text{Alg}:S_f,Q_f\rightarrow \widehat{Y}_Q$
		\STATE {\bfseries Initialize:} $\phi^{i=1:N_\text{e}}$
		\FOR{$i=1$ {\bfseries to} $N_\text{e}$}
		\STATE $\text{prev\_lid}=\widehat{\text{LID}}(\phi^i_0)$ 
		\STATE {\bfseries Initialize:} optimizer
		\FOR{$j=0$ {\bfseries to} MAX\_ITERATION}
		\STATE $\phi^i_{j+1} \leftarrow \phi^i_j - \nabla_{\phi^i_j}\mathcal{L}(\phi^i_j)$ from equation~\ref{equation:loss} or \ref{equation:semi}
		\STATE $\text{lid}=\widehat{\text{LID}}(\phi^i_{j+1})$
		\IF{lid $>$ prev\_lid}
		\STATE $\phi^i_* = \phi^i_{j + 1}$
		\STATE break
		\ENDIF
		\STATE $\text{prev\_lid}=\text{lid}$
		\ENDFOR
		\ENDFOR
		\STATE $S^\text{ESFR} = \{(z', y)|z'=\frac{1}{N_\text{e}}\sum_{i=1}^{N_\text{e}} g_{\phi^i_*}(z), (z,y)\in S_f\}$ \STATE $Q^\text{ESFR} = \{z'|z'=\frac{1}{N_\text{e}}\sum_{i=1}^{N_\text{e}} g_{\phi^i_*}(z), z\in Q_f\}$
		\STATE {\bfseries Output:}  $ \widehat{Y}_Q = \text{Alg}(S^\text{ESFR}, Q^\text{ESFR})$ 
	\end{algorithmic}
\end{algorithm}
\subsection{Proposed Methods}
\label{subsection:proposed}
\textbf{ESFR:}
We describe our unsupervised adaptation scheme for few-shot classification: Early-Stage Feature Reconstruction (ESFR).
Given a few-shot classification task and embedding network, we run feature reconstruction training by $\mathcal{L}_\text{FR}$ (\ref{equation:loss}) with initialized $\phi$.
At every training iteration, we measure the LID of $g_\phi$ with task samples by (\ref{equation:lid}) and we early stop the training when the LID starts to increase.
Task-adapted embeddings of $g_\phi(z)$ are used for the classification of the given few-shot task.
A wide range of metric-based and fine-tuning few-shot methods can be used with our embeddings for such classification.
To reduce the variance by random initial weights of $\phi$, we train $N_e$ reconstruction modules, separately, with different initial weights, and take the center of each sample's reconstructed embeddings.

\textbf{ESFR-Semi:}
We further investigate the semi-supervised version of our scheme, ESFR-Semi, by adding the support classification loss to the reconstruction loss (\ref{equation:loss}):
\begin{align}
\label{equation:semi}
&C^j(z,\phi,W,b) = \text{softmax}_j [Wg_\phi(z)+b] \notag \\
&\mathcal{L}_{\text{CE}}(\phi, W, b) = - \frac{1}{|S_f|}\sum_{(z_i,y_i)\in S_f} \log C^{y_i}(z_i, \phi, W, b) \notag \\
&\mathcal{L}_{\text{Semi}}(\phi, W, b) = \mathcal{L}_{\text{FR}}(\phi) + \lambda \mathcal{L}_{\text{CE}}(\phi, W, b)
\end{align}
where $\mathcal{L}_{\text{CE}}$ is the cross-entropy loss given by an affine classifier on new embeddings, and $\lambda$ is a trade-off parameter.
Additional weights $W$ and $b$ are jointly trained with $\phi$.
The trade-off parameter $\lambda$ is tuned using few-shot classification tasks from validation datasets as in \citet{LaplacianShot}.
Note that the only ESFR-Semi requires (while ESFR does not) few-shot classification task experiences to determine a certain parameter.

The overall methods are described in Algorithm~\ref{alg:proposed}.  \section{Related Work}

\textbf{Few-Shot Classification (FSC):}
Few-shot learning or few-shot classification methods have broad categories: optimization-based methods \cite{ravi2016optimization, MAML, LEO}, distance-based approaches \cite{vinyals2016matching, ProtoNet, tadam}, fine-tunings \cite{SimpleShot, tian2020rethinking, transBaseline}, etc.
A large portion of these methods is based on meta-learning \cite{metalearning}.
In meta-learning, training is done in a series of few-shot classification tasks (a.k.a. episodic training) to train the model in a way that reflects test-time scenarios.
Several recent studies \cite{SimpleShot, tian2020rethinking, transBaseline, LaplacianShot, TIM} have questioned the necessity of meta-learning on few-shot classification, reporting competitive performance on few-shot benchmarks without neither episodic training nor few-shot task experiences.
These methods solve the few-shot task by fine-tuning\footnote{Including the only change of class prototypes or last layer parameters.} a pre-trained embedding network trained on the base dataset with standard cross-entropy loss.
Our method lies in this line of research that doesn't require meta-learning. It seems possible to merge our method with meta-learning; we leave it as future work. 
\textbf{Unsupervised adaptation for FSC:}
We mentioned in the introduction that transductive methods, based on semi-supervised learning techniques, have been widely studied in few-shot classification.
In contrast, unsupervised adaptation for few-shot classification is an open field. To the best of our knowledge, we are the first to propose deep unsupervised adaptation for few-shot classification.
The closest related works are \citet{epnet} and \citet{TAFSSL}.
Embedding Propagation proposed by \citet{epnet} iteratively updates embeddings by a linear combination with the nearest neighbors' embeddings and improves few-shot classification performance.
\citet{TAFSSL} uses the principal component analysis to acquire major components from given task embeddings; this classical approach shows impressive performance gain.
However, these works are limited to an affine transformation of embeddings and do not benefit from deep learning; while our method finds non-linear patterns and correlations by the benefit of deep unsupervised learning.
Though unsupervised adaptation in few-shot classification had less attention, our experimental results show that it can outperform conventional transductive baselines, especially when extremely few labeled samples are available in the 1-shot setting. 
\textbf{Training behavior of DNNs:}
Our method is based on the property of Deep Neural Networks' (DNNs) training behavior that  \textit{DNNs learn to generalize before memorizing}.
The property is reported and studied in recent works \cite{ArpitL17, Lampinen19, stephenson21}, as mentioned in Section~\ref{subsection:feature_reconstruction}. 
Similar to our work, several methods of training DNNs with noisy labels are based on different training behaviors between generalization and memorization.
After the report of the property by \citet{ArpitL17}, several works \cite{hendrycks2019using, oymak2019generalization, song2020prestopping} suggest using early stopping to discard noisy information before memorization, \citet{jiang2018mentornet, yu2019does, sugiyama2018co} propose to gather the clean samples that exhibit high confident prediction in the early pattern learning for further training.
\citet{Ma18b} offer to use LID to detect and correct noisy label memorization, which strongly motivates our methodology.
However, these works are limited to the domain of learning from noisy labels and supervised learning.
In our work, we find the behavior also appears in unsupervised adaptation for few-shot classification.  \begin{table*}[t]
	\caption{Improvement by incorporating our method into baseline methods with ResNet-18/WRN-28-10 backbone on \textit{mini}-ImageNet and \textit{tiered}-ImageNet. $\dagger$ indicates the use of shifting-term (\ref{equation:shift}) during preprocessing.}
	\label{table:addition2}
	\begin{small}
		\begin{center}
			\begin{tabular}{cccccccccc}
				\hline
				& & \multicolumn{4}{c}{\textbf{ResNet-18}} & \multicolumn{4}{c}{\textbf{WRN-28-10}} \\
				& & \multicolumn{2}{c}{\textbf{\textit{mini}-ImageNet}} & \multicolumn{2}{c}{\textbf{\textit{tiered}-ImageNet}} & \multicolumn{2}{c}{\textbf{\textit{mini}-ImageNet}} & \multicolumn{2}{c}{\textbf{\textit{tiered}-ImageNet}} \\
				\multicolumn{2}{c}{\textbf{Method}} & \textbf{1-shot} & \textbf{5-shot} & \textbf{1-shot} & \textbf{5-shot}& \textbf{1-shot} & \textbf{5-shot}& \textbf{1-shot} & \textbf{5-shot} \\ \hline
				\multirow{2}{*}{(i)} & Linear & 62.45 & 79.32 & 68.49 & 83.77 & 64.53 & 80.81 & 69.78 & 84.91 \\
				&\cellcolor[HTML]{EFEFEF} + \textbf{ESFR} &\cellcolor[HTML]{EFEFEF} \textbf{70.38{\scriptsize+7.93}} &\cellcolor[HTML]{EFEFEF} \textbf{81.6{\scriptsize+2.28}} &\cellcolor[HTML]{EFEFEF} \textbf{76.98{\scriptsize+8.49}} &\cellcolor[HTML]{EFEFEF} \textbf{86.09{\scriptsize+2.32}} &\cellcolor[HTML]{EFEFEF} \textbf{73.33{\scriptsize+8.8}} &\cellcolor[HTML]{EFEFEF} \textbf{83.65{\scriptsize+2.84}} &\cellcolor[HTML]{EFEFEF} \textbf{78.57{\scriptsize+8.79}} &\cellcolor[HTML]{EFEFEF} \textbf{87.37{\scriptsize+2.46}} \\ \hline
				\multirow{2}{*}{(ii)} & NN & 64.04 & 79.71 & 71.60 & 84.62 & 66.73 & 81.85 & 72.97 & 85.74 \\
				&\cellcolor[HTML]{EFEFEF} + \textbf{ESFR} &\cellcolor[HTML]{EFEFEF} \textbf{70.94{\scriptsize+6.9}} &\cellcolor[HTML]{EFEFEF} \textbf{81.61{\scriptsize+1.9}} &\cellcolor[HTML]{EFEFEF} \textbf{77.44{\scriptsize+5.84}} &\cellcolor[HTML]{EFEFEF} \textbf{85.84{\scriptsize+1.22}} &\cellcolor[HTML]{EFEFEF} \textbf{74.01{\scriptsize+7.28}} &\cellcolor[HTML]{EFEFEF} \textbf{83.58{\scriptsize+1.73}} &\cellcolor[HTML]{EFEFEF} \textbf{79.13{\scriptsize+6.16}} &\cellcolor[HTML]{EFEFEF} \textbf{87.08{\scriptsize+1.34}} \\ \hline
				\multirow{2}{*}{(iii)} & CSPN$\dagger$ & 64.54 & 80.49 & 71.89 & 85.09 & 67.52 & 82.36 & 73.00 & 86.28 \\
				&\cellcolor[HTML]{EFEFEF} + \textbf{ESFR} &\cellcolor[HTML]{EFEFEF} \textbf{71.71{\scriptsize+7.17}} &\cellcolor[HTML]{EFEFEF} \textbf{82.22{\scriptsize+1.73}} &\cellcolor[HTML]{EFEFEF} \textbf{78.17{\scriptsize+6.28}} &\cellcolor[HTML]{EFEFEF} \textbf{86.38{\scriptsize+1.29}} &\cellcolor[HTML]{EFEFEF} \textbf{74.83{\scriptsize+7.31}} &\cellcolor[HTML]{EFEFEF} \textbf{84.17{\scriptsize+1.81}} &\cellcolor[HTML]{EFEFEF} \textbf{79.65{\scriptsize+6.65}} &\cellcolor[HTML]{EFEFEF} \textbf{87.57{\scriptsize+1.29}} \\ \hline
				\multirow{3}{*}{(iv)} & BD-CSPN$\dagger$ & 70.00 & \textbf{82.36} & 77.28 & \textbf{86.55} & 72.74 & 84.14 & 78.89 & \textbf{87.72} \\
				& \cellcolor[HTML]{EFEFEF} + \textbf{ESFR} & \cellcolor[HTML]{EFEFEF} \textbf{73.98{\scriptsize+3.98}} & \cellcolor[HTML]{EFEFEF} 82.32{\scriptsize-0.04} & \cellcolor[HTML]{EFEFEF} \textbf{80.13{\scriptsize+2.85}} & \cellcolor[HTML]{EFEFEF} 86.34{\scriptsize-0.21} & \cellcolor[HTML]{EFEFEF} \textbf{76.84{\scriptsize+4.10}} & \cellcolor[HTML]{EFEFEF} \textbf{84.36{\scriptsize+0.22}} & \cellcolor[HTML]{EFEFEF} \textbf{81.77{\scriptsize+2.88}} & \cellcolor[HTML]{EFEFEF} 87.61{\scriptsize-0.11} \\
				& \cellcolor[HTML]{EFEFEF} + \textbf{ESFR-Semi} & \cellcolor[HTML]{EFEFEF}  & \cellcolor[HTML]{EFEFEF} \textbf{82.89{\scriptsize+0.53}} & \cellcolor[HTML]{EFEFEF}  & \cellcolor[HTML]{EFEFEF} \textbf{86.83{\scriptsize+0.28}} & \cellcolor[HTML]{EFEFEF}  & \cellcolor[HTML]{EFEFEF} \textbf{84.97{\scriptsize+0.83}} & \cellcolor[HTML]{EFEFEF}  & \cellcolor[HTML]{EFEFEF} \textbf{88.10{\scriptsize+0.38}} \\ \hline
			\end{tabular}
		\end{center}
	\end{small}
\end{table*}

 \begin{table*}[t]
\begin{small}
		\caption{Comparison with state-of-the-art methods of 5-way 1- and 5-shot accuracy (in \%) on \textit{mini}-ImageNet, \textit{tiered}-ImageNet and CUB. The best results are reported in \textbf{bold}.}
		\label{table:sota}
		\begin{center}
			\begin{tabular}{lccccccc}
				& \textbf{}         & \multicolumn{2}{c}{\textbf{\textit{mini}-ImageNet}} & \multicolumn{2}{c}{\textbf{\textit{tiered}-ImageNet}} & \multicolumn{2}{c}{\textbf{CUB}}  \\
				\textbf{Method}                           & \textbf{Backbone} & \textbf{1-shot}      & \textbf{5-shot}     & \textbf{1-shot}       & \textbf{5-shot}      & \textbf{1-shot} & \textbf{5-shot} \\ \hline
				MAML \cite{MAML}                          & ResNet-18         & 49.61                & 65.72               & -                     & -                    & 68.42           & 83.47           \\
				Chen \cite{Chen19}                        & ResNet-18         & 51.87                & 75.68               & -                     & -                    & 67.02           & 83.58           \\
				ProtoNet \cite{ProtoNet}                  & ResNet-18         & 54.16                & 73.68               & -                     & -                    & 72.99           & 86.64           \\
				TPN \cite{TPN}                            & ResNet-12         & 59.46                & 75.65               & -                     & -                    & -               & -               \\
				TEAM \cite{Team}                          & ResNet-18         & 60.07                & 75.90               & -                     & -                    & 80.16           & 87.17           \\
				SimpleShot \cite{SimpleShot}              & ResNet-18         & 63.10                & 79.92               & 69.68                 & 84.56                & 70.28           & 86.37           \\
				CTM \cite{CTM}                            & ResNet-18         & 64.12                & 78.64               & 68.41                 & 84.28                & -               & -               \\
				FEAT \cite{FEAT}                          & ResNet-18         & 66.78                & 82.05               & 70.80                 & 84.79                & -               & -               \\
				BD-CSPN \cite{BDCSPN}                     & ResNet-18         & 70.00                & 82.36               & 77.28                 & 86.55                & 78.89           & 88.70               \\
				LaplacianShot \cite{LaplacianShot}        & ResNet-18         & 72.11                & 82.31               & 78.98                 & 86.39                & 80.96           & 88.68           \\
				\rowcolor[HTML]{EFEFEF} BD-CSPN + ESFR (Ours) & ResNet-18         & \textbf{73.98}       & 82.32      & \textbf{80.13}        & 86.34       & \textbf{82.68}  & 88.65  \\
				\rowcolor[HTML]{EFEFEF} BD-CSPN + ESFR-Semi (Ours) & ResNet-18         & -       & \textbf{82.89}      & -        & \textbf{86.83}       & -   & \textbf{89.10}  \\ \hline
LEO \cite{LEO}                            & WRN               & 61.76                & 77.59               & 66.33                 & 81.44                & -               & -               \\
				wDAE-GNN \cite{wDAE-GNN}                  & WRN               & 62.96                & 78.85               & 68.18                 & 83.09                & -               & -               \\
				FEAT \cite{FEAT}                          & WRN               & 65.10                & 81.11               & 70.41                 & 84.38                & -               & -               \\
				Tran. Baseline \cite{transBaseline}       & WRN               & 65.73                & 78.40               & 73.34                 & 85.50                & -               & -               \\
				SimpleShot \cite{SimpleShot}              & WRN               & 65.87                & 82.09               & 70.90                 & 85.76                & -               & -               \\
				SIB \cite{SIB}                            & WRN               & 70.0                 & 79.2                & -                     & -                    & -               & -               \\
				BD-CSPN \cite{BDCSPN}                     & WRN               & 72.74                & 84.14               & 78.89                 & 87.72                & -               & -               \\
				LaplacianShot \cite{LaplacianShot}        & WRN               & 74.86                & 84.13               & 80.18                 & 87.56                & -               & -               \\
				\rowcolor[HTML]{EFEFEF} BD-CSPN + ESFR (Ours) & WRN               & \textbf{76.84}       & \textbf{84.36}      & \textbf{81.77}        & 87.61       & -               & -               \\
				\rowcolor[HTML]{EFEFEF} BD-CSPN + ESFR-Semi (Ours) & WRN               & -       & \textbf{84.97}      & -        & \textbf{88.10}       & -               & -               \\ \hline
\end{tabular}
		\end{center}
	\end{small}
	\vskip -0.2in
\end{table*} \section{Experiments}
\subsection{Experimental Settings}
\textbf{Datasets:}
We evaluate our method on three standard datasets of few-shot classification: (1) \textit{mini}-ImageNet \cite{vinyals2016matching} dataset as in \citet{ravi2016optimization}, (2) \textit{tiered}-ImageNet dataset as in \citet{ren2018metalearning}, and (3) Caltech-UCSD Birds 200 (CUB) \cite{welinder2010caltech} as in \citet{Chen19}. Each dataset is divided into train/val/test splits according to references. All the images are resized to $84\times84$.

\textbf{Baseline Methods:}
We investigate three baseline few-shot classification methods (classifiers) used in conjunction with our method.
\begin{itemize}[leftmargin=*]
	\vspace{-0.1in}
	\item \textit{Linear}: Train an affine classifier on labeled support set with embeddings and use the trained classifier to classify the query samples.
	\item \textit{Nearest Neighbor} (NN): Compute the class prototype by the centroid of support sample embeddings in each class and classify the query sample to the class of the nearest prototype.\footnote{We use Euclidean as a distance metric.}
	\item \textit{BD-CSPN}: BD-CSPN \cite{BDCSPN} is chosen as a baseline transductive method.
	It can be used with pre-trained embeddings and achieves state-of-the-art performance on 5-shot (Table~\ref{table:sota}).
	To briefly summarize, BD-CSPN consists of two components: (1) \textit{shifting-term} for removing the cross-class bias between the support set and query set, (2) \textit{Prototype Rectification} (PR) that updates class prototypes by accounting pseudo-labeled query samples.
	Our particular interest is in PR since it is a baseline approach of transductive methods; a similar scheme was suggested in \citet{ren2018metalearning, CAN}.
\end{itemize}

\textbf{Evaluation Protocol:}
We evaluate our method on standard 5-way 1- and 5-shot settings with 15 query samples per class. We sampled 2,000 tasks for each experimental result. Our method uses the same fixed hyper-parameters for all experiments and settings. For the semi-supervised version of our method, with trade-off parameter $\lambda$, we tune $\lambda$ by selecting the best among $\lambda\in[0, 0.1, 0.2, 0.4, 0.8, 1.6]$ from 600 sampled few-shot tasks of validation dataset.

\subsection{Implementation Details}
\label{implementation_detail}
\textbf{Embedding Network:}
We follow the embedding network training procedure from \citet{LaplacianShot}, using backbone architectures: ResNet-18, WRN-28-10. We use \textit{from input to the average-pooled last residual block output} as an embedding network. Embedding networks are trained for 90 epochs, using stochastic gradient descent to minimize the standard cross-entropy on labeled base datasets. Label smoothing with parameter 0.1 is used for more general features. Random cropping, color jittering, and random horizontal flipping are applied for data augmentation.\footnote{Data augmentations are used only for embedding network pre-training.} The initial learning rate is 0.1 and shrank by $\frac{1}{10}$ at 45 and 66 epochs. Mini-batch sizes of 256 and 128 are used for ResNet-18 and WRN-28-10 training, respectively. The best performing embedding network on the 5-way 1-shot task of validation split dataset is chosen; while using the nearest neighbor classifier and l2-normalized embeddings for classification.

\textbf{Reconstruction Training:}
For the reconstruction module, we use 4-layer fully connected deep neural network with ReLU activation. Each layer has the size of units equal to the embedding dimension, and weight initialization follows the TensorFlow \cite{abadi2016tensorflow} default setting (GlorotUniform). As an optimizer, we use Adam \cite{kingma2014adam} with a default learning rate of 1e-3. For dropout \cite{dropout}, we use the maximum possible rate of $0.5$. Finally, we take the ensemble of $N_e=5$ reconstruction modules.

\textbf{Preprocessing:}
\citet{SimpleShot} reports the importance of preprocessing in few-shot classification. We apply centering and l2-normalization to embedding network's output for reconstruction training and baseline methods. For centering, we subtract the center of task sample embeddings from each embedding as in \citet{TAFSSL} since it performs better on all baseline methods. For BD-CSPN \cite{BDCSPN}, the aforementioned shifting-term defined as:
\begin{equation}
	\label{equation:shift}
	\triangle=\frac{1}{|S|}\sum_{x_s\in S}f(x_s) - \frac{1}{|Q|}\sum_{x_q\in Q}f(x_q)
\end{equation}
is added for query sample before centering as in \citet{LaplacianShot, BDCSPN}. For reconstruction training, when computing the reconstruction loss, output embeddings of the reconstruction module and the pre-trained embedding network follows the same preprocessing. For few-shot classification with new embeddings, we apply only l2-normalization since it performs the best.\footnote{Note that applying only l2-normalization for baseline methods worsen the performance.} We refer to Appendix A for more details on preprocessing. \begin{table*}[t]
	\caption{Ablation study evaluating the effects of embedding ensemble and dropout perturbation.}
	\label{table:ablation}
	\begin{small}
		\begin{center}
			\begin{tabular}{c|cc|cc|cc}
				\hline
				& \multicolumn{2}{c|}{\textbf{\textit{mini}-ImageNet}} & \multicolumn{2}{c|}{\textbf{\textit{tiered}-ImageNet}} & \multicolumn{2}{c|}{\textbf{CUB}}          \\
				\textbf{Method}                                         & \textbf{1-shot}      & \textbf{5-shot}      & \textbf{1-shot}       & \textbf{5-shot}       & \textbf{1-shot} & \textbf{5-shot} \\ \hline
				NN with ResNet-18                                     & 64.04                & 79.71                & 71.60                 & 84.62                 & 71.43           & 86.44           \\ \hline
				(i) w/o Dropout and Ensemble                        & 66.87                & 80.59                & 73.39                 & 84.60                 & 75.46           & 87.02           \\
				(ii) w/o Ensemble                                       & 69.66                & 81.10                & 76.31                 & 85.33                 & 78.32           & 87.63           \\
				(iii) w/o Dropout                                   & 68.90                & 81.53                & 75.39                 & 85.31                 & 77.32          & 87.66           \\
				\rowcolor[HTML]{EFEFEF} \textbf{NN + ESFR} & \textbf{70.94}       & \textbf{81.61}       & \textbf{77.44}        & \textbf{85.84}        & \textbf{79.44}  & \textbf{88.02}  \\ \hline
			\end{tabular}
		\end{center}
	\end{small}
	\vskip -0.2in
\end{table*} \subsection{Results}
\label{subsection:main_results}
\textbf{Improvement by ESFR:}
We validate the improvement of our method when used in conjunction with commonly used few-shot classification methods. We evaluate our method in the most common 5-way 1- and 5-shot settings on standard \textit{mini}-ImageNet, \textit{tiered}-ImageNet datasets with ResNet-18, WRN-28-10 backbone networks. The results are listed in Table~\ref{table:addition2}.

In Table~\ref{table:addition2}-(i, ii), we investigate our method with linear and NN classifiers. Each is a widely used baseline method in unsupervised representation learning and few-shot classification. Our method provides noticeable improvements in all settings;  +5.9\%$\sim$8.9\% for 1-shot and +1.5\%$\sim$3.1\% for 5-shot settings. This roughly indicates that our embedding adaptation provides well-clustered new embeddings; hence samples can be classified by simple affine classifiers.

In Table~\ref{table:addition2}-(iii, iv), we compare our method with the semi-supervised learning approach, Prototype-Rectification (PR) \cite{BDCSPN}. For a fair comparison, we use the same preprocessing with shifting-term (\ref{equation:shift}) and cosine similarity-based nearest neighbor (CSPN) classifier as in BD-CSPN$\dagger$ (which includes PR); the result is described as CSPN$\dagger$ + ESFR. We can observe that ESFR further improves accuracies on 1-shot by +0.8\%$\sim$2.1\% compared to BD-CSPN$\dagger$ while showing comparable improvements on 5-shot. This result implies that, though unsupervised adaptation has received less attention in few-shot classification, well-designed unsupervised learning can provide comparable improvements to conventional transductive methods.

Performance can be further improved by incorporating our method into BD-CSPN, a transductive method. The result is described in Table~\ref{table:addition2}-(iv). For the 5-shot setting, there was no additional gain by combining our method; however, for the 1-shot setting, improvements of +2.9\%$\sim$4.1\% indicates that ESFR can offer a complementary improvement to PR.

In the last row of Table~\ref{table:addition2}-(iv), we also investigate ESFR-semi that uses the semi-supervised loss $\mathcal{L}_\text{Semi}$ (\ref{equation:semi}) for embedding adaptation. As mentioned in Section~\ref{implementation_detail}, we chose the trade-off parameter $\lambda$ that best performs with validation tasks. For the 1-shot setting, we find that $\lambda=0$ performs the best that does not use label information during adaptation. For the 5-shot setting, ESFR-semi gives an additional +0.3\%$\sim$0.8\% gain over BD-CSPN.

\textbf{Comparison with prior work:}
We compare our method with prior few-shot classification methods\footnote{We separately compared the work by \cite{TIM} in Appendix B; since they use strong prior that the number of query samples per class is equal.} on \textit{mini}-ImageNet, \textit{tiered}-ImageNet, and CUB datasets in Table~\ref{table:sota}. We use BD-CSPN + ESFR and BD-CSPN + ESFR-Semi as our methods.

For the 1-shot setting, our method achieves new state-of-the-art performance on all datasets and backbone networks. To be specific, our method outperforms the previous state-of-the-art LaplacianShot \cite{LaplacianShot} by +1.2\%$\sim$2.0\%. We note that our implementation is based on LaplacianShot; BD-CSPN + ESFR shares many aspects including the baseline method BD-CSPN and embedding network pre-training.

For the 5-shot setting, our method shows comparable performance to the state-of-the-art. Note that in Table~\ref{table:addition2}, CSPN + ESFR, which does not benefit from transductive inference, shows subequal performance with  BD-CSPN + ESFR. For BD-CSPN + ESFR-Semi, our method provides further improvements of +0.3\%$\sim$0.8\%.

\textbf{Ablation study:}
We conduct an ablation analysis on the effects of different components of the proposed method. We use a NN classifier with ResNet-18 backbone network for these experiments; our experimental results are on \textit{mini}-ImageNet, \textit{tiered}-ImageNet, and CUB with 5-way 1- and 5-shot tasks. Table~\ref{table:ablation} shows the influences of the embedding ensemble and dropout perturbation.

In Section~\ref{subsection:proposed}, we proposed the embedding ensemble to reduce the variance by random initialization. The empirical result in Table~\ref{table:ablation}-(iii) shows that the ensemble consistently provides improvements in all settings. In Section~\ref{subsection:feature_reconstruction}, we expected the effectiveness of dropout perturbation since noise perturbations tend to make memorization harder. The result in Table~\ref{table:ablation}-(ii) shows consistent improvements by dropout; supports our expectation. Finally, we observe that the complete version with both the embedding ensemble and dropout perturbation outperforms the other configurations.

\textbf{Comparison with prior embedding adaptation}:
We compare our method with prior embedding adaptation methods that are based on affine transformations: 1) Embedding Propagation (EP) from \citet{epnet}, 2) Principal Component Analysis (PCA)- and Independent Component Analysis (ICA)\footnote{\citet{TAFSSL} explained their method as Independent Component Analysis (ICA), but it seems the difference between their PCA- and ICA-based methods is only in the whitening; for ICA, they whiten by the projection of $1^Tz=0$.} -based methods from \citet{TAFSSL}. For a fair comparison, we experiment with the same nearest neighbor classifier and pre-trained embeddings. We use the released official code of each method. Table~\ref{Table:PCA} describes the results of 1- and 5-shot settings on \textit{mini}-ImageNet with WRN backbone.
\begin{table}[ht]
	\vskip -0.15in
	\caption{Comparison with prior embedding adaptation methods}
	\label{Table:PCA}
	\begin{small}
		\begin{center}
			\begin{tabular}{lcc}
				\hline
				& \textbf{1-shot} & \textbf{5-shot} \\ \hline
				NN		       & 66.73 $\pm$ 0.44          & 81.85 $\pm$ 0.31          \\
				NN + PCA       & 69.63 $\pm$ 0.50          & 82.28 $\pm$ 0.32          \\
				NN + EP        & 70.58 $\pm$ 0.47          & 82.73 $\pm$ 0.30          \\
				NN + ICA       & 72.19 $\pm$ 0.54          & 83.12 $\pm$ 0.32          \\
				\rowcolor[HTML]{EFEFEF} 
				\textbf{NN + ESFR (Ours)} & \textbf{74.01 $\pm$ 0.51} & \textbf{83.58 $\pm$ 0.31}         \\ \hline
			\end{tabular}
		\end{center}
	\end{small}
	\vskip -0.15in
\end{table}

Our method outperforms both \citet{epnet} and \citet{TAFSSL} on all settings. Extracting shared or correlated features reminds us of the concept of PCA; however, compared to PCA, our method advantage from discovering non-linear patterns and correlates.  \section{Conclusion}
We propose the novel unsupervised embedding adaptation scheme based on the finding that \textit{deep neural networks learn to generalize before memorizing}.
Our method, ESFR, provides task-adapted embeddings of generalizable features by feature reconstruction training and LID-based early stopping.
Experimental results show that well-designed unsupervised adaptation can consistently improve baseline methods; outperform conventional transductive methods; be further improved by joint usage with a transductive method.
ESFR used in conjunction with the transductive method achieves new state-of-the-art performance on the 1-shot setting.
We hope that our work will become a starting point for future unsupervised learning studies on few-shot classification. 
\section*{Acknowledgements}
This work was supported by the National Research Foundation of Korea (NRF) grant funded by the Korea government (MSIT) [2021R1A2C2007518].

\bibliography{ref}
\bibliographystyle{icml2021}

\renewcommand\thesection{\Alph{section}}
\onecolumn
\setcounter{section}{0}
\icmltitle{Appendix: Unsupervised Embedding Adaptation via Early-Stage Feature Reconstruction for Few-Shot Classification}

\section{Preprocessing}
In this section, we describe the preprocessing including equations in our paper.
Assume we are given the embedding support set $S_f$ and embedding query set $Q_f$.
We apply centering and l2-normalization to the embedding samples for reconstruction training as described in (\ref{preprocessing}).
Preprocessed embeddings $z\in Z_\text{preprocess}$ are used as an input to the reconstruction module $g_\phi$.
The same preprocessing (centering and l2-normalization) is applied at the output of reconstruction module to compute the reconstruction loss $\mathcal{L}_\text{FR}$ as in (\ref{preprocessing2}).
\begin{align}
&\bar{z} = \frac{1}{|S_f\cup Q_f|} \sum_{z\in S_f\cup Q_f} z \\
&S_\text{preprocessed} = \Big\{(z', y)|z'=\frac{z-\bar{z}}{{\Vert z-\bar{z} \Vert}_2}, (z, y)\in S_f\Big\}, \quad Q_\text{preprocessed} = \Big\{z'|z'=\frac{z-\bar{z}}{{\Vert z-\bar{z} \Vert}_2}, z\in Q_f\Big\} \label{preprocessing}\\
&Z_\text{preprocessed} = S_\text{preprocessed} \cup Q_\text{preprocessed} \\
&\bar{z}_\phi = \frac{1}{|Z_\text{preprocessed}|}\sum_{z\in Z_\text{preprocessed}} \mathbb{E}_\mu [g_\phi(z\odot\mu)] \\
&\mathcal{L}_\text{FR}(\phi) = - \frac{1}{|Z_\text{preprocessed}|}\sum_{z\in Z_\text{preprocessed}} \mathbb{E}_\mu \Big[z^T \frac{g_\phi(z\odot \mu) - \bar{z}_\phi}{{\Vert g_\phi(z\odot \mu) - \bar{z}_\phi \Vert}_2}\Big] \label{preprocessing2}
\end{align}

As new embeddings for few-shot classification, we apply only l2-normalization since it performs the best.
The new embedding sets $S^\text{ESFR}$ and $Q^\text{ESFR}$ for the few-shot classification task are as follows:
\begin{align}
& S^\text{ESFR} = \Big\{(z',y)|z'=\frac{1}{N_e}\sum_{i=1}^{N_e}\frac{g_{\phi_*^i}(z)}{{\Vert g_{\phi_*^i}(z) \Vert}_2}, (z,y)\in S_\text{preprocessed} \Big\} \\
& Q^\text{ESFR} = \Big\{z'|z'=\frac{1}{N_e}\sum_{i=1}^{N_e}\frac{g_{\phi_*^i}(z)}{{\Vert g_{\phi_*^i}(z) \Vert}_2}, z\in Q_\text{preprocessed} \Big\}
\end{align}

For BD-CSPN \cite{BDCSPN} and our method used with BD-CSPN, additional shifting-term is added for query samples before preprocessing.
In this case, we define $S_\text{preprocess}$ and $Q_\text{preprocess}$ as follows:
\begin{align}
& \text{shifting-term: } \triangle = \frac{1}{S_f}\sum_{z_s\in S_f} z_s  - \frac{1}{Q_f}\sum_{z_q\in Q_f}z_q \\
& Q_f^\text{shifted} = \Big\{z'|z'=z + \triangle, z\in Q_f \Big\} \\
&\bar{z} = \frac{1}{|S_f\cup Q_f^\text{shifted}|} \sum_{z\in S_f\cup Q_f^\text{shifted}} z \\
&S_\text{preprocessed} = \Big\{(z', y)|z'=\frac{z-\bar{z}}{{\Vert z-\bar{z} \Vert}_2}, (z, y)\in S_f\Big\}, \quad Q_\text{preprocessed} = \Big\{z'|z'=\frac{z-\bar{z}}{{\Vert z-\bar{z} \Vert}_2}, z\in Q_f^\text{shifted}\Big\}
\end{align}

\newpage
\section{Comparison to TIM}
\begin{table*}[h]
	\vskip -0.2in
	\begin{small}
		\caption{Table describes the performance comparison with TIM \cite{TIM} of 5-way 1- and 5-shot accuracies (in \%) on \textit{mini}-ImageNet, \textit{tiered}-ImageNet and CUB. The performance of TIM-GD is from the paper \cite{TIM}. We use preprocessing with shifting-term to acquire new embeddings for TIM-GD + ESFR since it performs better.}
		\label{table:tim}
		\begin{center}
			\begin{tabular}{lccccccc}
				& \textbf{}         & \multicolumn{2}{c}{\textbf{\textit{mini}-ImageNet}} & \multicolumn{2}{c}{\textbf{\textit{tiered}-ImageNet}} & \multicolumn{2}{c}{\textbf{CUB}}  \\
				\textbf{Method}                           & \textbf{Backbone} & \textbf{1-shot}      & \textbf{5-shot}     & \textbf{1-shot}       & \textbf{5-shot}      & \textbf{1-shot} & \textbf{5-shot} \\ \hline
				TIM-GD \cite{TIM}                         & ResNet-18         & 73.9                 & \textbf{85.0}       & 79.9                  & \textbf{88.5}        & 82.2            & \textbf{90.8}   \\
				\rowcolor[HTML]{EFEFEF} TIM-GD + ESFR & ResNet-18         & \textbf{76.02}       & 84.42               & \textbf{82.03}        & 88.11                & \textbf{84.69}  & 90.43           \\ \hline
				TIM-GD \cite{TIM}                         & WRN               & 77.8                 & \textbf{87.4}       & 82.1                  & \textbf{89.8}        & -               & -               \\
				\rowcolor[HTML]{EFEFEF} TIM-GD + ESFR & WRN               & \textbf{79.25}       & 86.38               & \textbf{83.58}        & 89.44                & -               & -               \\ \hline
			\end{tabular}
		\end{center}
	\end{small}
	\vskip -0.1in
\end{table*}

We separately compare the performance of our method with TIM \cite{TIM} since we believe TIM uses a strong prior that query samples per class are balanced in the standard few-shot classification benchmarks (e.g., 15-query samples per class); while our method combined with the baseline methods (NN, Linear, BD-CSPN \cite{BDCSPN}) does not utilize any query statistics.
We find that TIM's proposed regularization term with conditional entropy and label-marginal entropy forces balancing among the predicted number of query samples per class.
To be specific, the conditional entropy minimization term encourages the classification model to output confident prediction and the label-marginal entropy maximization term encourages marginal predicted label distribution to be uniform.
When both conditional and label-marginal entropy terms are used simultaneously, the predicted labels close to one-hot from uniform class distribution, resulting in the same number of predicted samples for each class.
This seems helpful in the balanced query class distribution setting where all query samples per class are the same, but its possible use case will be limited.
We find that query class imbalance setting can easily ruin TIM's performance in Section~\ref{section:imbalance}.

To investigate our method when using query statistics, we experiment with TIM-GD + ESFR.
Table~\ref{table:tim} shows the results on standard mini-ImageNet, tiered-ImageNet, and CUB with 5-way 1- and 5-shot settings.
For 1-shot settings, our method improves the performance of TIM by 1.5\%$\sim$2.5\% across all datasets and backbones.
As mentioned before (in Section~5.3), this indicates that our method can offer a complementary improvement to semi-supervised learning techniques such as TIM for 1-shot.
For 5-shot settings, our method decreases the performance by 0.4\%$\sim$1.0\%.
The decrease in 5-shot performance encourages further research about the simultaneous use of pseudo-label information and unsupervised information, which we leave as future work.

\section{Ablation: noise level of dropout}
\begin{table}[h]
	\vspace{-0.15in}
	\caption{The table shows the influence of drop-rate applied to our method. We experimented with ResNet-18 backbone on \textit{mini}-ImageNet and \textit{tiered}-ImageNet.}
	\label{table:noise}
	\begin{small}
		\begin{center}
			\begin{tabular}{ccccc}
				\hline
				\textbf{} & \multicolumn{2}{c}{\textbf{\textit{mini}-ImageNet}} & \multicolumn{2}{c}{\textbf{\textit{tiered}-ImageNet}} \\
				\textbf{rate} & \textbf{1shot} & \textbf{5shot} & \textbf{1shot} & \textbf{5shot} \\ \hline
				0. & 68.90 & 81.53 & 75.39 & 85.31 \\
				0.1 & 69.41 & 81.59 & 75.90 & 85.50 \\
				0.2 & 69.90 & 81.70 & 76.39 & 85.63 \\
				0.3 & 70.39 & \textbf{81.71} & 76.78 & 85.71 \\
				0.4 & 70.63 & \textbf{81.71} & 77.23 & 85.77 \\
				\rowcolor[HTML]{EFEFEF} 0.5 & \textbf{70.94} & 81.61 & \textbf{77.44} & \textbf{85.84} \\ \hline
			\end{tabular}
		\end{center}
	\end{small}
\end{table}
We further investigate the effect of the dropout noise level.
In the main text, we argued that multiplicative noise by dropout seems well suited for our method. Experiments in Table~\ref{table:noise} with various drop-rate show that the dropout can be used in our method without careful tuning.

\section{Few-shot classification with imbalance query class distribution}
\label{section:imbalance}
\begin{table}[t]
	\begin{small}
		\caption{
			This table shows few-shot classification performance when the numbers of query samples per class are imbalanced.
			For standard settings, the number of query samples per class is equally 15, given as (15, 15, 15, 15, 15).
			For the imbalance case, we set the number of query samples per class as (11, 13, 15, 17, 19) and (7, 11, 15, 19, 23).
			The $\pm$ describes 95\% confidence interval.
			For these results, we use our implementation version of TIM-GD \cite{TIM}, which matches the original paper's performance.
			For BD-CSPN \cite{BDCSPN} with an imbalance number of query samples, we do not use shift-term since it worsens the performance.}
		\label{table:imbalance}
		\begin{center}
			\begin{tabular}{cccccccc}
				\hline
				\multicolumn{8}{c}{\textbf{\textit{mini}-ImageNet}} \\ \hline
				\multirow{2}{*}{\textbf{Method}} & \multirow{2}{*}{\textbf{Backbone}} & \multicolumn{2}{c}{\textbf{(15, 15, 15, 15, 15)}} & \multicolumn{2}{c}{\textbf{(11, 13, 15, 17, 19)}} & \multicolumn{2}{c}{\textbf{(7, 11, 15, 19, 23)}} \\
				&  & \textbf{1-shot} & \textbf{5-shot} & \textbf{1-shot} & \textbf{5-shot} & \textbf{1-shot} & \textbf{5-shot} \\ \hline
				TIM-GD & ResNet-18 & 73.67$\pm$0.33 & 85.01$\pm$0.19 & 68.93$\pm$0.30 & 79.05$\pm$0.17 & 66.04$\pm$0.28 & 75.60$\pm$0.16 \\ \hline
				NN & ResNet-18 & 64.04$\pm$0.44 & 79.71$\pm$0.32 & 63.73$\pm$0.46 & 80.01$\pm$0.33 & 63.25$\pm$0.47 & 79.88$\pm$0.33 \\
				\rowcolor[HTML]{EFEFEF}+ESFR & ResNet-18 & 70.94$\pm$0.50 & 81.61$\pm$0.33 & 70.32$\pm$0.52 & 81.35$\pm$0.33 & 69.74$\pm$0.53 & 81.12$\pm$0.34 \\ \hline
				BD-CSPN & ResNet-18 & 70.00$\pm$0.51 & 82.36$\pm$0.32 & 68.99$\pm$0.51 & 81.49$\pm$0.34 & 68.26$\pm$0.52 & 81.12$\pm$0.34 \\
				\rowcolor[HTML]{EFEFEF}+ESFR & ResNet-18 & 73.98$\pm$0.55 & 82.32$\pm$0.33 & 72.39$\pm$0.56 & 81.51$\pm$0.34 & 71.74$\pm$0.57 & 81.17$\pm$0.35 \\ \hline
				TIM & WRN & 77.60$\pm$0.31 & 87.31$\pm$0.17 & 72.03$\pm$0.28 & 80.91$\pm$0.16 & 68.86$\pm$0.26 & 77.28$\pm$0.15 \\ \hline
				NN & WRN & 66.73$\pm$0.44 & 81.85$\pm$0.31 & 66.64$\pm$0.46 & 82.07$\pm$0.31 & 66.30$\pm$0.47 & 81.98$\pm$0.32 \\
				\rowcolor[HTML]{EFEFEF}+ESFR & WRN & 74.01$\pm$0.51 & 83.58$\pm$0.31 & 73.34$\pm$0.51 & 83.27$\pm$0.32 & 72.89$\pm$0.52 & 83.03$\pm$0.33 \\ \hline
				BD-CSPN & WRN & 72.74$\pm$0.49 & 84.14$\pm$0.30 & 71.67$\pm$0.51 & 83.34$\pm$0.32 & 71.19$\pm$0.51 & 83.02$\pm$0.33 \\
				\rowcolor[HTML]{EFEFEF}+ESFR & WRN & 76.84$\pm$0.54 & 84.36$\pm$0.32 & 75.26$\pm$0.55 & 83.48$\pm$0.33 & 74.66$\pm$0.55 & 83.09$\pm$0.34 \\ \hline

				\multicolumn{8}{c}{\textbf{\textit{tiered}-ImageNet}} \\ \hline
				\multirow{2}{*}{\textbf{Method}} & \multirow{2}{*}{\textbf{Backbone}} & \multicolumn{2}{c}{\textbf{(15, 15, 15, 15, 15)}} & \multicolumn{2}{c}{\textbf{(11, 13, 15, 17, 19)}} & \multicolumn{2}{c}{\textbf{(7, 11, 15, 19, 23)}} \\
				&  & \textbf{1-shot} & \textbf{5-shot} & \textbf{1-shot} & \textbf{5-shot} & \textbf{1-shot} & \textbf{5-shot} \\ \hline
				TIM-GD & ResNet-18 & 79.99$\pm$0.33 & 88.62$\pm$0.20 & 74.21$\pm$0.29 & 81.93$\pm$0.18 & 70.95$\pm$0.28 & 78.36$\pm$0.17 \\ \hline
				NN & ResNet-18 & 71.60$\pm$0.49 & 84.62$\pm$0.36 & 71.10$\pm$0.49 & 84.59$\pm$0.35 & 70.51$\pm$0.49 & 84.52$\pm$0.35 \\
				\rowcolor[HTML]{EFEFEF}+ESFR & ResNet-18 & 77.44$\pm$0.52 & 85.84$\pm$0.35 & 76.77$\pm$0.53 & 85.64$\pm$0.35 & 76.21$\pm$0.54 & 85.42$\pm$0.36 \\ \hline
				BD-CSPN & ResNet-18 & 77.28$\pm$0.52 & 86.55$\pm$0.34 & 76.38$\pm$0.52 & 85.89$\pm$0.35 & 75.63$\pm$0.53 & 85.65$\pm$0.36 \\
				\rowcolor[HTML]{EFEFEF}+ESFR & ResNet-18 & 80.13$\pm$0.56 & 86.34$\pm$0.36 & 78.72$\pm$0.57 & 85.76$\pm$0.36 & 78.12$\pm$0.57 & 85.50$\pm$0.37 \\ \hline
				TIM & WRN & 82.18$\pm$0.32 & 89.87$\pm$0.19 & 76.11$\pm$0.28 & 83.18$\pm$0.17 & 72.72$\pm$0.27 & 79.55$\pm$0.16 \\ \hline
				NN & WRN & 72.97$\pm$0.49 & 85.74$\pm$0.34 & 72.17$\pm$0.48 & 85.79$\pm$0.34 & 71.57$\pm$0.49 & 85.70$\pm$0.34 \\
				\rowcolor[HTML]{EFEFEF}+ESFR & WRN & 79.13$\pm$0.52 & 87.08$\pm$0.34 & 78.30$\pm$0.53 & 86.90$\pm$0.34 & 77.67$\pm$0.53 & 86.69$\pm$0.34 \\ \hline
				BD-CSPN & WRN & 78.89$\pm$0.52 & 87.72$\pm$0.32 & 77.71$\pm$0.52 & 87.11$\pm$0.34 & 77.05$\pm$0.53 & 86.86$\pm$0.35 \\
				\rowcolor[HTML]{EFEFEF}+ESFR & WRN & 81.77$\pm$0.55 & 87.61$\pm$0.34 & 80.50$\pm$0.55 & 87.04$\pm$0.35 & 79.67$\pm$0.56 & 86.72$\pm$0.35 \\ \hline
			\end{tabular}
		\end{center}
	\end{small}
\end{table}
To verify our method's robustness on various query settings, we experiment with the setting when the numbers of query samples per class are imbalanced.
We set the number of query samples per class as (11, 13, 15, 17, 19) and (7, 11, 15, 19, 23).
Table~\ref{table:imbalance} shows that our method consistently improves the performance of few-shot classification regardless of the query imbalance setting.
To be more specific, the improvement by our method in different query settings varies within $<1.5\%$; thus, our method is robust to different query settings.
In contrast, TIM \cite{TIM} that uses the strong prior about query statistics, suffers from the change in query setting.
The performance of TIM on the 5-shot when the number of query samples per class is (7, 11, 15, 19, 23) shows $-$6.2\%$\sim$$-$4.2\% performance decrease compare to the baseline (NN).
\section{Naively applied unsupervised learning methods}
\begin{table}[h]
	\caption{Experiment settings}
	\label{table:settings}
	\begin{small}
		\begin{center}
			\begin{tabular}{cc}
				& Candidates \\ \hline
				\multicolumn{1}{c|}{Learning rate} & 1e-3, \textbf{1e-4} \\ \hline
				\multicolumn{1}{c|}{Classifier} & Linear, \textbf{Cosine} \\ \hline
				\multicolumn{1}{c|}{Additional module} & \textbf{2-layer FCN}, None \\ \hline
				\multicolumn{1}{c|}{\begin{tabular}[c]{@{}c@{}}update weights \\ of embedding networks\end{tabular}} & \begin{tabular}[c]{@{}c@{}}None, All,\\ \textbf{only-the-last-residual-block}\end{tabular} \\ \hline
				\multicolumn{1}{c|}{New embeddings} & \textbf{Backbone output}, Additional module output \\ \hline
			\end{tabular}
		\end{center}
	\end{small}
\end{table}
We experiment if naively applied unsupervised (or self-supervised) learning can improve few-shot classification in the standard settings.
For a fair comparison, we use the pre-trained embeddings of ResNet-18 on \textit{mini}-ImageNet.
We test with pretext task-based self-supervised methods of rotation \cite{Rotnet} and jigsaw \cite{jigsaw}.
For both methods, we use grid search to find the best performing settings; shown in Table~\ref{table:settings}.
An additional module is inserted between the embedding network and classifier and we use hidden dimensions from \citet{Su2020When}.
For jigsaw tasks, we use 35-permutations from \citet{Su2020When}.
For both methods, the same setting with the bold font on Table~~\ref{table:settings} performs the best.

\begin{table}[h]
	\caption{Naively applied unsupervised learning results}
	\label{table:results}
	\begin{small}
		\begin{center}
			\begin{tabular}{ccc}
				\textbf{Method} & \multicolumn{2}{c}{\textbf{\textit{mini}-ImageNet 1-shot accuracy}} \\ \hline
				\textbf{NN} & \multicolumn{2}{c}{64.04} \\
				\textbf{+ESFR} & \multicolumn{2}{c}{70.94 {\scriptsize+6.90}} \\ \hline
				& \textbf{(1) Converged} & \textbf{(2) Oracle early stopping} \\ \hline
				\textbf{Jigsaw} & 33.1 & 67.22 {\scriptsize+3.18} \\
				\textbf{Rotation} & 32.2 & 66.70 {\scriptsize+2.66} \\ \hline
			\end{tabular}
		\end{center}
	\end{small}
\end{table}
Table-\ref{table:results} shows the results with (1) new embeddings are provided when training becomes converged and (2) new embeddings are given via oracle early stopping at the best performing training iteration (for $\geq1$).
The result with converged embeddings shows that the naively applied self-supervised learning fails to improve few-shot classification performance.
Note that our method achieves $70.94$ on the same setting.
Our method outperforms both the rotation- and jigsaw-based\footnote{We improved the performance with the jigsaw task after the rebuttal period. Originally performance with rotation task performs better.} unsupervised learning methods that even contain oracle early-stopping.

\section{Experiments with Conv4}
\begin{table}[h]
\label{table:conv4}
	\begin{small}
		\caption{The table shows the experimental results of our method with Conv4-64 backbone on \textit{mini}-ImageNet and \textit{tiered}-ImageNet.}
		\begin{center}
			\begin{tabular}{ccccc}
				\hline
				\textbf{} & \multicolumn{2}{c}{\textbf{mini-ImageNet}} & \multicolumn{2}{c}{\textbf{tiered-ImageNet}} \\
				\textbf{Method} & \textbf{1-shot} & \textbf{5-shot} & \textbf{1-shot} & \textbf{5-shot} \\ \hline
				NN & 50.72 & 67.17 & 52.18 & 69.60 \\
				\rowcolor[HTML]{EFEFEF} 
				\textbf{+ ESFR} & \textbf{54.63 {\scriptsize+3.91}} & \textbf{68.32 {\scriptsize+1.15}} & \textbf{57.56 {\scriptsize+5.38}} & \textbf{71.46 {\scriptsize+1.86}} \\ \hline
				BD-CSPN & 52.73 & 68.5 & 54.94 & 71.53 \\
				\rowcolor[HTML]{EFEFEF} 
				\textbf{+ ESFR} & \textbf{56.24 {\scriptsize+3.51}} & \textbf{69.18 {\scriptsize+0.68}} & \textbf{60.16 {\scriptsize+5.22}} & \textbf{72.37 {\scriptsize+0.84}} \\ \hline
			\end{tabular}
		\end{center}
	\end{small}
\end{table}
We use pre-trained Conv4-64 backbones following the settings of \citet{SimpleShot}.
We applied the same preprocessing strategy as in the main text.
For the reconstruction module, we find that the bottleneck structure (Section~3.1) is helpful for Conv4-64; while reconstructed embeddings still outperform the encoded ones.
Thus, we use 800-400-800-1600 as hidden dimensions.

Table-\ref{table:conv4} shows experimental results with Conv4-64.
As in the experimental results with ResNet and WideResNet, our method consistently improves the performance of the baseline methods: NN and BD-CSPN.
ESFR also offers a complementary improvement to BD-CSPN \cite{BDCSPN} in 1-shot settings.
Compare to prior state-of-the-art methods with Conv4-64, our method with BD-CSPN has slightly lower performance.
\footnote{\citet{SIB} shows $58.0\%$ and $70.7\%$ accuracies on {\it mini}-ImageNet in 1- and 5-shot settings, respectively.} \end{document}